%% file: acl_latex.tex
\pdfoutput=1

\documentclass[11pt]{article}

\usepackage[]{acl}

\usepackage{times}
\usepackage{latexsym}

\usepackage[T1]{fontenc}

\usepackage[utf8]{inputenc}

\usepackage{microtype}
\usepackage{amsfonts}
\usepackage{amssymb}
\usepackage{multirow}
\usepackage{adjustbox}
\usepackage[english]{babel} % American English
\usepackage{arydshln}
\usepackage[ruled,linesnumbered]{algorithm2e}
\usepackage{color}
\usepackage{booktabs}
\usepackage{CJKutf8}
\usepackage{color}
\usepackage{amsmath}
\usepackage{graphicx}
\usepackage{tikz}
\usepackage{pgfplots}

\title{\textsc{GL-CLeF}: A Global--Local Contrastive Learning Framework for Cross-lingual Spoken Language Understanding}

\author{ Libo Qin$^{1}$\thanks{\ \  Work done during internship at Microsoft Research Asia and remote visiting at National University of Singapore.}, Qiguang Chen$^{1}$, Tianbao Xie$^{1}$, Qixin Li$^1$, \\ \textbf{Jian-Guang Lou$^2$ , Wanxiang Che$^1$\thanks{\ \  Email corresponding.}, Min-Yen Kan$^3$}
	\\$^1$Research Center for Social Computing and Information Retrieval\\
	Harbin Institute of Technology, China \\
	$^2$ Microsoft Research Asia, Beijing, China\\
	$^3$Department of Computer Science, National University of Singapore\\
	\{lbqin,tianbaoxie,qxli,car\}@ir.hit.edu.cn;
	jlou@microsoft.com; kanmy@comp.nus.edu.sg}

\begin{document}
	\maketitle
	\begin{CJK*}{UTF8}{gbsn}
		\begin{abstract}
			Due to high data demands of current methods, attention to zero-shot cross-lingual spoken language understanding (SLU) has grown, as such approaches 
			greatly reduce human annotation effort.
			However, existing models solely rely on shared parameters, which can only perform implicit alignment across languages. We present \textbf{G}lobal--\textbf{L}ocal \textbf{C}ontrastive \textbf{\textsc{Le}}arning \textbf{F}ramework (\textsc{GL-CLeF}) to address this shortcoming.
			Specifically, we employ contrastive learning, leveraging bilingual dictionaries to construct multilingual views of the same 
			utterance, then encourage their representations to be more similar than negative example pairs, which achieves to explicitly aligned representations of similar sentences across languages.
			In addition, a key step in \textsc{GL-CLeF} is a proposed \texttt{Local} and \texttt{Global} component, which achieves a fine-grained cross-lingual transfer (i.e., \textit{sentence-level} \texttt{Local} intent transfer, \textit{token-level} \texttt{Local} slot transfer, and \textit{semantic-level} \texttt{Global} transfer across intent and slot).
			Experiments on MultiATIS++ show that \textsc{GL-CLeF} achieves the best performance and successfully pulls representations of similar sentences across languages closer.
		\end{abstract}
		
		\section{Introduction}
		\label{Introduction}
		\input{Introduction.tex}
		
		\section{Background}
		\label{Background}
		\input{Background.tex}
		
		\section{Model}
		\label{Model}
		\input{Model.tex}

		\section{Experiments}
		\label{experiments}
		\input{Experiments.tex}

		\section{Related Work}
		\label{related-work}
		\input{related-work.tex}
		
		\section{Conclusion}
		\label{conclusion}
		\input{conclusion.tex}

		\bibliography{anthology,custom}
		\bibliographystyle{acl_natbib}
	\end{CJK*}
\end{document}

%% file: Introduction.tex
Spoken language understanding (SLU) is a critical component in task-oriented dialogue systems~\cite{tur2011spoken,ijcai2021-622}. 
It usually includes two sub-tasks: intent detection to identify users’ intents and slot filling to extract semantic constituents from the user's query.
With the advent of deep neural network methods, SLU has met with remarkable success.
However, existing SLU models rely on large amounts of annotated data, which makes it hard to scale to low-resource languages that lack large amounts of labeled data.
To address this shortcoming, zero-shot cross-lingual SLU generalization leverages the labeled training data in high-resource languages to transfer the trained model to a target, low-resource language, which gains increasing attention.

\begin{figure}[t]
	\centering
	\includegraphics[width=0.45\textwidth]{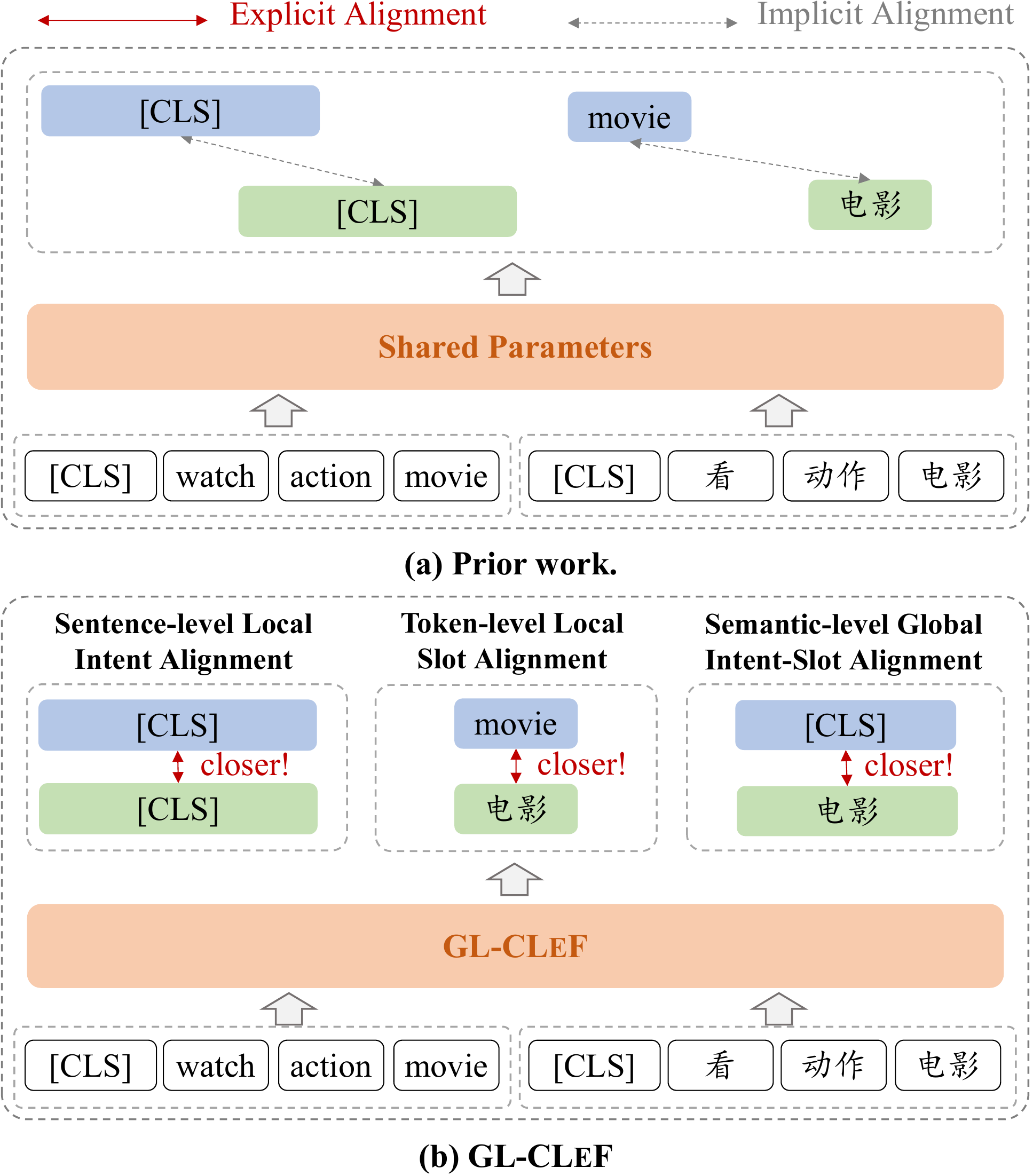}
	\caption{(a) Prior work (Implicit Alignment); (b) \textsc{GL-CLeF} (Explicit Alignment). 
	Different color denotes representations across different languages. \texttt{[CLS]} represents the sentence representation.
	}
	\label{fig:comparison}
\end{figure}

To this end, many works have been explored for zero-shot cross-lingual SLU.
Multilingual BERT (mBERT)~\cite{devlin-etal-2019-bert}, a  cross-lingual contextual pre-trained model from a large amount of multi-lingual corpus multi-lingual corpus, has achieved considerable performance for zero-shot cross-lingual SLU. 
\citet{liu2020attention} further build an attention-informed mixed-language training by generating bi-lingual code-switched data to implicitly align keywords (e.g., slots) between source and target language.
\citet{ijcai2020-533} extend the idea to a multilingual code-switched setting, aligning the source language to multiple target languages.
This approach currently achieves the state-of-the-art performance for zero-shot cross-lingual SLU.
Though achieving promising performance, as shown in Figure~\ref{fig:comparison} (a), the above methods solely rely on shared parameters and can only perform implicit alignment across languages, which brings two challenges.
First, such implicit alignment process seems to be a black box, which not only seriously affects the alignment representation but also makes it hard to analyze the alignment mechanism.
Second, prior work do not distinguish between the varying granularities of the tasks: the intent detection is \textit{sentence-level} and the slot filling is \textit{token-level}, which does not offer fine-grained cross-lingual transfer for \textit{token-level} slot filling.

To solve the aforementioned challenges, we propose a \textbf{G}lobal--\textbf{L}ocal \textbf{C}ontrastive \textbf{\textsc{Le}}arning \textbf{F}ramework (\textsc{GL-CLeF}) for zero-shot cross-lingual SLU.
For the first challenge, as shown in Figure~\ref{fig:comparison} (b), the key insight in \textsc{GL-CLeF} is to explicitly ensure that representations of similar sentences across languages are pulled closer together via contrastive learning (CL).
Specifically, we leverage bilingual dictionaries to generate multi-lingual code-switched data pairs, which can be regarded as cross-lingual views with the same meaning.
With the use of CL, our model is able to learn to distinguish the code-switched utterance of an input sentence from a set of negative examples, and thus encourages representations of similar sentences between source language and target language closer.

For the second challenge, SLU requires accomplishing tasks at two different levels: \textit{token-level} slot filling and \textit{sentence-level} intent detection.  As such, simply leveraging ordinary {\it sentence-level} contrastive learning is ineffective for fine-grained knowledge transfer in {\it token-level} slot filling.
Therefore, we first introduce a \texttt{Local} module in \textsc{GL-CLeF} to learn different granularity alignment representations (i.e., \textit{sentence-level} \texttt{Local} intent CL and \textit{token-level} \texttt{local} slot CL).
To be specific, \textit{sentence-level} \texttt{Local} intent CL and \textit{token-level} \texttt{local} slot CL are introduced for aligning similar sentence and token representations across different languages for intent detection and slot filling, respectively. 
In addition, we further argue that slot and intent are highly correlated and have similar semantic meanings in a sentence.  This phenomenon can serve as a signal for self-supervised alignment across intent and slots.
Therefore, a \texttt{Global} module named \textit{semantic-level} \texttt{global} intent--slot CL is further proposed to bring the representations of slot and intents within a sentence closer together.

We conduct experiments on MultiATIS++ \cite{xu-etal-2020-end}, which includes nine different languages.
Our experiments show that \textsc{GL-CLeF} achieves state-of-the-art results of 54.09\% sentence accuracy, outperforming the previous best by 10.06\% on average.
Besides, extensive analysis experiments demonstrate that \textsc{GL-CLeF} has successfully reduced the representation gap between different languages.

To facilitate further research,
codes are publicly available at \url{https://github.com/LightChen233/GL-CLeF}.

%% file: Background.tex
\begin{figure}[t]
	\centering
	\includegraphics[width=0.47\textwidth]{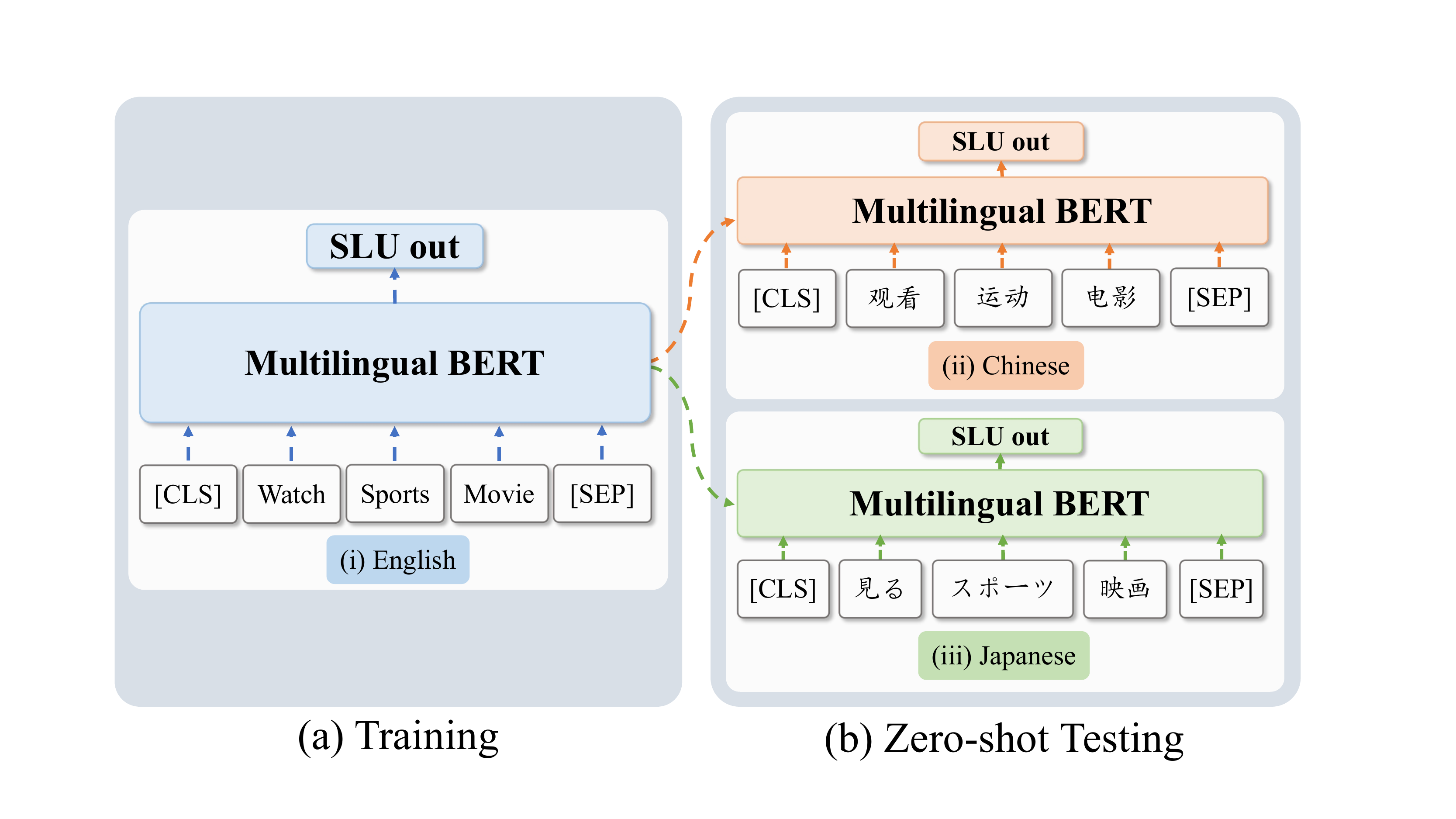}
	\caption{Zero-shot cross-lingual SLU.}
	\label{fig:Zero-shot}
\end{figure}
\begin{figure*}[t]
	\centering
	\includegraphics[width=0.95\textwidth]{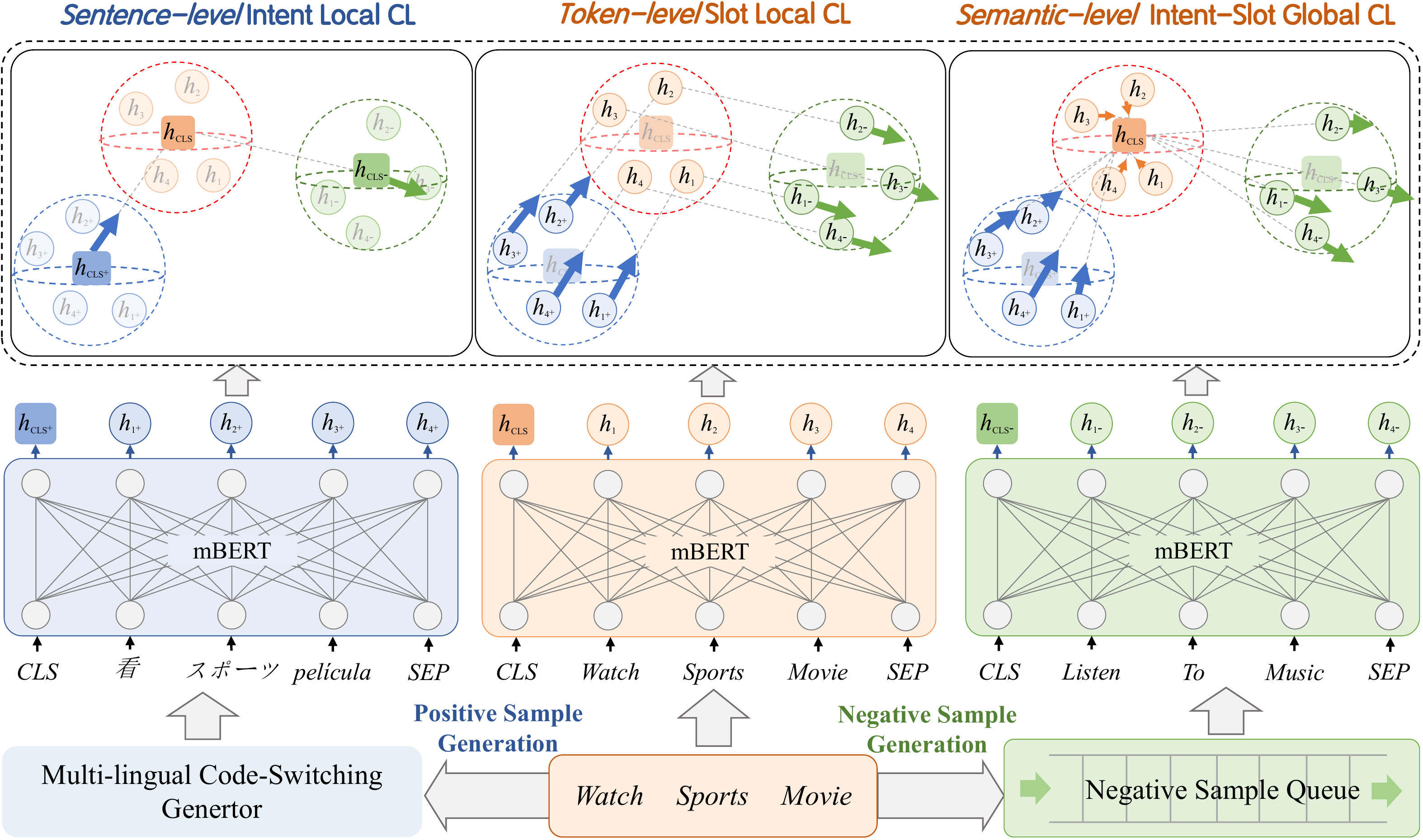}
	\caption{The main architecture of \textsc{GL-CLeF}. 
	The boxes shown in figures are each sentence representation, while the circles are token representations.
	The dash lines and arrows in the top of the pictures on boxes and circles represent the direction of pushing in different levels made by contrastive learning. 
	Different color denotes different representation spaces against anchor utterance, positive samples and negative samples. For simplicity, we only draw one case of \textit{token-level} slot CL.
	} 
	\label{fig:framework}
\end{figure*}
We first describe traditional SLU before the specifics of zero-shot cross-lingual version of SLU.

\paragraph{Traditional SLU in Task-oriented Dialogue.}
SLU in Task-oriented Dialogue contains two sub-tasks: {\it Intent Detection} and {\it Slot Filling}.

\noindent $\cdot$ {\it Intent Detection:} Given input utterance $\mathbf{x}$, this is a classification problem to decide the corresponding intent label $o^{I}$. \vspace{2mm}

\noindent $\cdot$ {\it Slot Filling:} Often modeled as a sequence labeling task that maps an input word sequence $\mathbf{x} = (x_{1},\dots, x_{n})$ to slots sequence  $\mathbf{o}^{S} = (o^{S}_{1},\dots, o^{S}_{n})$, where $n$ denotes the length of sentence $\mathbf{x}$. \vspace{2mm}

Since the two tasks of intent detection and slot filling are highly correlated, it is common to adopt a joint model that can capture shared knowledge.  We follow the formalism from \newcite{goo-etal-2018-slot}, formulated as 
$(o^{I}, o^{S}) = f(\mathbf{x})$, where $f$ is the trained model.

\paragraph{Zero-shot Cross-lingual SLU.}
This means that a SLU model is trained in a source language, e.g., English ({\it cf.} Figure~\ref{fig:Zero-shot} (a)) and directly applied to other target languages  ({\it cf.} Figure~\ref{fig:Zero-shot} (b)).

Formally, given each instance $\mathbf{x}_{tgt}$ in a target language, the model $f$ which is trained on the source language is directly used for predicting its intent and slots:
\begin{eqnarray}
	(o_{tgt}^{I}, o_{tgt}^{S}) = f (\mathbf{x}_{tgt}),
\end{eqnarray}
where $tgt$ represents the target language.

%% file: Model.tex
We describe the general approach to general SLU task first, before describing our \textsc{GL-CLeF} model which explicitly uses contrastive learning to explicitly achieve cross-lingual alignment.
The main architecture of  \textsc{GL-CLeF} is illustrated in Figure~\ref{fig:framework}.

\subsection{A Generic SLU model}

\paragraph{Encoder.} Given each input utterance $x = (x_{1}, x_{2}, \dots, x_{n})$, the input sequence can be constructed by adding specific tokens  $\textbf{x} = ({\texttt{[CLS]}, x_{1}, x_{2}, ..., x_{n}, \texttt{[SEP]}})$, where $\texttt{[CLS]}$ denotes the special symbol for representing the whole sequence, and $\texttt{[SEP]}$ can be used for separating non-consecutive token sequences \cite{devlin-etal-2019-bert}.
Then, we follow \newcite{ijcai2020-533} to first generate multi-lingual code-switched data. Then, we employ mBERT model to take code-switched data for encoding their representations $\mathbf{H}$ =  ($\boldsymbol{h}_{\texttt{CLS}}$, $\boldsymbol{h}_{1}$, \dots , $\boldsymbol{h}_{n}$, $\boldsymbol{h}_{\texttt{SEP}}$).

\paragraph{Slot Filling.} 

Since mBERT produces subword-resolution embeddings, we follow \newcite{wang-etal-2019-cross} and adopt the first sub-token's representation as the whole word representation and use the hidden state to predict each slot:
$\boldsymbol{o}^{S}_t = \operatorname{softmax}({ \boldsymbol{W}}^s \boldsymbol{h}_t + \boldsymbol{b}^s) \,, $
\noindent  where \(\boldsymbol{h}_t\) denotes the first sub-token representation of word \(x_t\); \(\boldsymbol{W}^s\) and \(\boldsymbol{b}^s\) refer to the trainable parameters.

\paragraph{Intent Detection.}
We input the sentence representation $\boldsymbol{h}_\texttt{CLS}$ to a classification layer to find the label $o^{I}$:
$o^{I} =\operatorname{softmax}(\boldsymbol{W^I}{\boldsymbol{h}}_{\texttt{CLS}} + \boldsymbol{b^I})$, 
where $\boldsymbol{W^I}$ and $\boldsymbol{b^{I}}$ are tuneable parameters. 

\subsection{Global--local Contrastive Learning Framework}
We introduce our global--local contrastive learning framework (\textsc{GL-CLeF}) in detail, which consists of three modules: 1) a \textit{sentence-level} \texttt{local} intent contrastive learning (CL) module to align sentence representation across languages for intent detection, 2) a \textit{token-level} \texttt{local} slot CL module to align token representations across languages for slot filling, and 3) \textit{semantic-level} \texttt{global} intent--slot CL to align representations between a slot and an intent.

\subsubsection{Positive and Negative Samples Construction}
For contrastive learning, the key operation is to choose appropriate positive and negative pairs against to the original (anchor) utterance.
\paragraph{Positive Samples.}
Positive samples should preserve the same semantics compared against the anchor utterance.
Therefore, given each anchor utterance $\textbf{x} = ({\texttt{[CLS]}, x_{1}, x_{2}, ..., x_{n}, \texttt{[SEP]}})$, we follow \newcite{ijcai2020-533} to use bilingual dictionaries \cite{lample2018word} to generate multi-lingual code-switched data, which is considered as the positive samples $\textbf{x}_{+}$.
Specifically, for each word $x_{t}$ in $\textbf{x}$, $x_{t}$ is randomly chosen to be replaced with a translation provisioned from a bilingual dictionary to generate a positive sample.
For example, given an anchor utterance \textit{``watch sports movie''} in English, we can generate a positive multi-lingual code-switched sample \textit{``看 (watch/zh) スポーツ (sports/ja) película (movie/es)''}  ({\it cf.} Figure~\ref{fig:framework}).  Such a pair of anchor utterance and multi-lingual code-switched sample can be regarded as cross-lingual views of the same meaning across different languages.
$\textbf{x}_{+}$ is fed into mBERT to obtain the corresponding representations $\mathbf{H}_{+}$ =  ($\boldsymbol{h}_{\texttt{CLS}^{+}}$, $\boldsymbol{h}_{1^{+}}$, \dots , $\boldsymbol{h}_{n^{+}}$, $\boldsymbol{h}_{\texttt{SEP}^{+}}$).

\paragraph{Negative Samples.} A natural approach for generating negative samples is randomly choosing other queries in a batch.
However, this method requires the recoding of the negative samples, hurting efficiency.
Inspired by \newcite{he2020momentum}, in \textsc{GL-CLeF}, we maintain a negative sample queue, where the previously encoded original anchor utterance $\textbf{x}$, positive samples $\textbf{x}_+$ and previous negative samples $\textbf{x}_-$ are also progressively reused as negative samples.
This enables us to reuse the encoded samples from the immediate preceding batches, so as to eliminate the unnecessary negative encoding process.
The negative sample queues for \texttt{[CLS]} and sentence representation are represented as:
$\mathbf{H}_{\texttt{CLS}^-}$=$\{\boldsymbol{h}_{\texttt{CLS}^-}^k\}_{k=0}^{K-1}$, $\mathbf{H}_{S^-}$=$\{\mathbf{H}_{S^-}^k\}_{k=0}^{K-1}$
, where K is the maximum capacity for negative queue.

\subsubsection{\texttt{Local} Module}

\paragraph{\textit{Sentence-level} \texttt{Local} Intent CL.}
Since intent detection is a \textit{sentence-level} classification task, aligning sentence representation across languages is the goal of zero-shot cross-lingual intent detection task.
Therefore, in \textsc{GL-CLeF}, we propose a \textit{sentence-level} \texttt{local} intent CL loss to explicitly encourage the model to align similar sentence representations into the same local space across languages for intent detection.  Formally, this is formulated as:

\begin{small}
    \begin{gather*}
    	\mathcal{L}_{\text{LI}}\text{=}\!-\!\text{log} \frac{s(h_{\text{CLS}} , h_{\text{CLS}^{\text{+}}})}{s(h_{\text{CLS}}, h_{\text{CLS}^{\text{+}}})+\sum_{\text{k=0}}^{\text{K-1}} \!s(h_{\text{CLS}},h_{\text{CLS}^\text{-}}^{k})},
    \end{gather*}
\end{small}
where $s(p,q)$ denotes the dot product between $p$ and $q$; $\tau$ is a scalar temperature parameter.

\paragraph{\textit{Token-level} \texttt{Local} Slot CL.}

As slot filling is a token-level task, we propose a \textit{token-level} \texttt{local} slot CL loss to help the model to consider token alignment for slot filling, achieving fine-grained cross-lingual transfer. We apply toke-level CL for all tokens in the query. Now, we calculate the $i$th token CL loss for simplicity:

\begin{small}
	\begin{gather*}
		\mathcal{L}^i_{\text{LS}}\text{=}-\sum^n_{j=1}\log \frac{s(h_{\text{i}},h_{\text{j}^{\text{+}}})}{s(h_{\text{i}},h_{\text{j}^{\text{+}}})+\sum_{k\text{=}0}^{\text{K-1}}s(h_{\text{i}},h_{\text{j}^\text{-}}^{k})}/n,
	\end{gather*}
\end{small}
\noindent where the final $	\mathcal{L}_{\text{LS}}$ is the summation of all tokens CL loss.

\subsubsection{\texttt{Global} Module}
\paragraph{\textit{Semantic-level} \texttt{Global} Intent-slot CL.}

We noted that slots and intent are often highly related semantically when they belong to the same query.
Therefore, we think that the intent in a sentence and its own slots can naturally constitute a form of positive pairings, and the corresponding slots in other sentences can form negative pairs.
We thus further introduce a \textit{semantic-level} \texttt{global} intent--slot CL loss to model the semantic interaction between slots and intent, which may further improve cross-lingual transfer between them.  Formally:

\begin{small}
	\begin{gather*}
	    \mathcal{L}_{\text{GIS1}}\! =\! \text{-}\sum^n_{j=1}\log \frac{s(h_{\text{CLS}},h_{\text{j}})}{s(h_{\text{CLS}},h_{\text{j}})+\sum_{k\text{=}0}^{\text{K-1}}s(h_{\text{CLS}},h_{\text{j}^\text{-}}^{k})}/n,\\
		\mathcal{L}_{\text{GIS2}}\! =\! \text{-}\sum^n_{j=1}\log \frac{s(h_{\text{CLS}},h_{\text{j}^{\text{+}}})}{s(h_{\text{CLS}},h_{\text{j}^{\text{+}}})+\sum_{k\text{=}0}^{\text{K-1}}s(h_{\text{CLS}},h_{\text{j}^\text{-}}^{k})}/n,\\
		\mathcal{L}_{\text{GIS}} =\mathcal{L}_{\text{GIS1}}+\mathcal{L}_{\text{GIS2}},
	\end{gather*}
\end{small}

\noindent where we consider CL loss from both anchor sentences ($\mathcal{L}_{\text{GIS1}}$) and code-switched sentence ($\mathcal{L}_{\text{GIS2}}$), and add them to do semantic-level contrastive learning ($\mathcal{L}_{\text{GIS}}$) .

\subsection{Training}
\subsubsection{Intent Detection Loss}
\begin{equation}
\mathcal { L } _ { I } \triangleq -\sum _ { i = 1 } ^ { n_{I} }  \hat { {\bf{y}} } _ { i} ^ {I } \log \left( {\bf{o}} _ {i } ^ { I } \right),
\end{equation}
where ${\hat { {\bf{y}} } _ { i} ^ { I } }$ are the gold intent label and $n_{I}$ is the number of intent labels.
\subsubsection{Slot Filling Loss}
\begin{equation}
\mathcal { L } _ { S} \triangleq - \sum _ { j = 1 } ^ { n } \sum _ { i = 1 } ^ { n_{S} } { \hat { {\bf{y}} } _ {  j} ^ { i, S } \log \left( {\bf{y}} _ {  j} ^ { i,S } \right)},
\end{equation}
where ${\hat { {\bf{y}} } _ { j} ^ { i, S} }$ are the gold slot label for $j$th token; $n_{S}$ is the number of slot labels.
\subsubsection{Overall Loss}
The overall objective in \textsc{GL-CLeF} is a tuned linear combination of the individual losses:
\begin{footnotesize}
	\begin{equation}
		\mathcal{L}=\lambda_I \mathcal{L}_{I}+\lambda_S \mathcal{L}_{S}+\lambda_{\text{LI}} \mathcal{L}_{\text{LI}}+\lambda_{\text{LS}} \mathcal{L}_{\text{LS}}+\lambda_{\text{GIS}} \mathcal{L}_{\text{GIS}},
	\end{equation}
\end{footnotesize}
\noindent where $\lambda_*$ are tuning parameters for each loss component.

%% file: Experiments.tex
We use the latest multilingual benchmark dataset of MultiATIS++~\cite{xu-etal-2020-end} which consists of 9 languages including English (en), Spanish (es), Portuguese (pt), German (de), French (fr), Chinese (zh), Japanese (ja), Hindi (hi), and Turkish (tr).

\begin{table*}
	\centering
	\begin{adjustbox}{width=\textwidth}
		\begin{tabular}{l|ccccccccc|c}
			\hline
			\textbf{Intent Accuracy} & en & de & es & fr & hi & ja & pt & tr & zh & \textbf{AVG} \\ 
			\hline
			mBERT*~\cite{xu-etal-2020-end} & - & 95.27 & 96.35 & 95.92 & 80.96 & 79.42 & 94.96 & 69.59 & 86.27 & -\\
			mBERT$^\dagger$~\cite{devlin-etal-2019-bert} & 98.54 & 95.40 & 96.30 & 94.31 & 82.41 & 76.18 & 94.95 & 75.10 & 82.53 & 88.42\\
			Ensemble-Net*~\cite{razumovskaia2021crossing} & 90.26 & 92.50 & 96.64 & 95.18 & 77.88 & 77.04 & 95.30 & 75.04 & 84.99 & 87.20 \\
			CoSDA$^\dagger$~\cite{ijcai2020-533} & 95.74 & 94.06 & 92.29 & 77.04 & 82.75 & 73.25 & 93.05 & 80.42 & 78.95 & 87.32 \\
			\hline
			\textsc{GL-CLeF} & \textbf{98.77} & \textbf{97.53} & \textbf{97.05} & \textbf{97.72} & \textbf{86.00} & \textbf{82.84} & \textbf{96.08} & \textbf{83.92} & \textbf{87.68} & \textbf{91.95} \\
			\hline
			\hline
			\textbf{Slot F1} & en & de & es & fr & hi & ja & pt & tr & zh & \textbf{AVG} \\ 
			\hline
			Ensemble-Net*~\cite{razumovskaia2021crossing} & 85.05 & 82.75 & 77.56 & 76.19 & 14.14 & 9.44 & 74.00 & 45.63 & 37.29 & 55.78\\
			mBERT*~\cite{xu-etal-2020-end} & - & 82.61 & 74.98 & 75.71 & 31.21 & 35.75 & 74.05 & 23.75 & 62.27 & -\\
			mBERT$^\dagger$~\cite{devlin-etal-2019-bert} & 95.11 & 80.11 & 78.22 & 82.25 & 26.71 & 25.40 & 72.37 & 41.49 & 53.22 & 61.66 \\
			CoSDA$^\dagger$~\cite{ijcai2020-533} & 92.29 & 81.37 & 76.94 & 79.36 & 64.06 & 66.62 & 75.05 & 48.77 & 77.32 & 73.47 \\
			\hline
			\textsc{GL-CLeF} & \textbf{95.39} & \textbf{86.30} & \textbf{85.22} & \textbf{84.31} & \textbf{70.34} & \textbf{73.12} & \textbf{81.83} & \textbf{65.85} & \textbf{77.61} & \textbf{80.00} \\
			
			\hline
			\hline
			\textbf{Overall Accuracy} & en & de & es & fr & hi & ja & pt & tr & zh & \textbf{AVG} \\ 
			\hline
			AR-S2S-PTR*~\cite{zhu2020don} & 86.83 & 34.00 & 40.72 & 17.22 & 7.45 & 10.04 & 33.38 & -- & 23.74 & -\\
			IT-S2S-PTR*~\cite{zhu2020don} & 87.23 & 39.46 & 50.06 & 46.78 & 11.42 & 12.60 & 39.30 & -- & 28.72 & -\\
				mBERT$^\dagger$~\cite{devlin-etal-2019-bert} & 87.12 & 52.69 & 52.02 & 37.29 & 4.92 & 7.11 & 43.49 & 4.33 & 18.58 & 36.29 \\
			CoSDA$^\dagger$~\cite{ijcai2020-533} & 77.04 & 57.06 & 46.62 & 50.06 & 26.20 & 28.89 & 48.77 & 15.24 & 46.36 & 44.03 \\
			\hline
			\textsc{GL-CLeF} & \textbf{88.02} & \textbf{66.03} & \textbf{59.53} & \textbf{57.02} & \textbf{34.83} & \textbf{41.42} & \textbf{60.43} & \textbf{28.95} & \textbf{50.62} & \textbf{54.09} \\
			\hline
		\end{tabular}
	\end{adjustbox}
	\caption{Results on MultiATIS++. We report both individual and average (AVG) test results on slot filling, intent detection accuracy, and overall accuracy. Results with ``*'' are taken from the corresponding published paper, while results with $^\dagger$ are obtained by re-implemented. '--' denotes missing results from the published work.} \label{table:result}
\end{table*}
\subsection{Experimental Setting}
We use the base case multilingual BERT (mBERT), which has $N = 12$ attention heads and $M = 12$ transformer blocks.
We select the best hyperparameters by searching a combination of batch size, learning rate with the following ranges: learning rate $\{2\times 10^{-7},5\times 10^{-7},  1\times 10^{-6},  2\times 10^{-6},5\times 10^{-6},6\times 10^{-6},5\times 10^{-5},5\times 10^{-4}\}$; batch size $\{4,8,16,32\}$; max size of negative queue $\{4,8,16,32\}$;
For all experiments, we select the best-performing model over the dev set and evaluate on test datasets. 
All experiments are conducted at TITAN XP and V100.
\subsection{Baselines}
To verify the effect of \textsc{GL-CLeF}, we compare our model with the following state-of-the-art baselines:\\
1) {\texttt{mBERT.}} mBERT\footnote{https://github.com/google-research/bert/blob/master/multilingual.md} follows the same model architecture and training procedure as BERT~\cite{devlin-etal-2019-bert}, but trains on the Wikipedia pages of 104 languages with a shared subword vocabulary.  This allows mBERT to share embeddings across languages, which achieves promising performance on various cross-lingual NLP tasks;\\
2) {\texttt{Ensemble-Net.}} \citet{razumovskaia2021crossing} propose an \texttt{Ensemble-Net} where predictions are determined by 8 independent models through majority voting, each separately
trained on a single source language, which achieves promising performance on zero-shot cross-lingual SLU;\\
3) {\texttt{AR-S2S-PTR.}} \citet{2020} proposed a unified sequence-to-sequence models with pointer generator network for cross-lingual SLU;\\
4) {\texttt{IT-S2S-PTR.}} \citet{zhu2020don} proposed a non-autoregressive parser based on the insertion transformer. It speeds up decoding and gain improvements in cross-lingual SLU transfer;\\
5) {\texttt{CoSDA.}} \citet{ijcai2020-533} propose a data augmentation framework to generate multi-lingual code-switching data to fine-tune mBERT, which encourages the model to align representations from source and multiple target languages.

\subsection{Main Results}
Following ~\newcite{goo-etal-2018-slot},
we evaluate the performance of slot filling using F1
score, intent prediction using accuracy, and the sentence-level semantic frame parsing
using overall accuracy which represents all metrics
are right in an utterance.

From the results in Table~\ref{table:result}, we observe that: (1) \texttt{CoSDA} achieves better performance than no alignment work \texttt{mBERT} and even outperforms the \texttt{Ensemble-Net}.
This is because that such implicit alignment does align representations to some extent, compared against \texttt{mBERT}.
(2) Our framework achieves the state-of-the art performance and beats \texttt{CoSDA} with 10.06\% average improvements on overall accuracy.
This demonstrates that \texttt{\textsc{GL-CLeF}} explicitly pull similar representations across languages closer, which outperforms the implicit alignment manner.

\subsection{Analysis}
To understand \textsc{GL-CLeF} in more depth, we 
perform comprehensive studies to answer the following research questions (RQs): \\
(1) Do the \texttt{local} intent and slot CLs benefit \textit{sentence-} and \textit{token-level} representation alignment? 
(2) Can \textit{semantic-level} \texttt{global} intent-slot CL boost the overall sentence accuracy?
(3) Are \texttt{local} intent CL and \texttt{local} slot CL complementary? 
(4) Does \textsc{GL-CLeF} pull similar representations across languages closer? 
(5) Does \textsc{GL-CLeF} improve over other pre-trained models? 
(6) Does \textsc{GL-CLeF} generalize to non pre-trained models? 
(7) Is \textsc{GL-CLeF} robust to the one-to-many translation problem?
\paragraph{Answer 1: \texttt{Local} intent CL and slot CL align similar sentence and token representations across languages.}
We investigate the effect of the \texttt{local} intent CL and \texttt{local} slot CL mechanism, by removing the \texttt{local} intent CL and slot CL, respectively (Figure \ref{fig:ablation}, ``-- LI'' and ``-- LS'' (Col~1,2)).
For the effectiveness of \texttt{local} intent CL, we find the performance of intent detection averaged on 9 languages drops by 3.52\% against the full system ({\it ibid.} final, RHS column).
This is because \textit{sentence-level} intent CL loss can pull sentence representations closer across languages.

Similarly, considering the effectiveness of \texttt{local} slot CL, we find the performance of slot filling averaged on 9 languages drops by 2.44\% against the full system. We attribute performance drops  to the fact that \texttt{local} slot CL successfully make a fine-grained cross-lingual knowledge transfer for aligning token representation across languages, which is essential for \textit{token-level} cross-lingual slot filling tasks.

\begin{table*}
	\centering
	\begin{adjustbox}{width=\textwidth}
		\begin{tabular}{l|ccccccccc|c}
			\hline
			\textbf{Intent Accuracy} & en & de & es & fr & hi & ja & pt & tr & zh & \textbf{AVG} \\ 
			\hline
			BiLSTM~\cite{hochreiter1997long} & 72.56 & 70.96 & 70.35 & 60.05 & 64.50 & 64.33 & 71.75 & 56.22 & 60.13 & 65.65\\
			BiLSTM+$\textsc{GL-CLeF}$ & \textbf{84.77} & \textbf{74.44} & \textbf{71.09} & \textbf{69.53} & \textbf{65.29} & \textbf{66.14} & \textbf{77.02} & \textbf{63.36} & \textbf{67.08} & \textbf{70.97}\\
			\hdashline
			XLM-R~\cite{ conneau-etal-2020-unsupervised} & 98.32 & 97.19 & 98.03 & 94.94 & 88.91 & 88.50 & 96.41 & 72.45 & 91.15 & 93.02\\
			XLM-R+$\textsc{GL-CLeF}$ & \textbf{98.66} & \textbf{98.43} & \textbf{98.04} & \textbf{97.85} & \textbf{93.84} & \textbf{88.83} & \textbf{97.76} & \textbf{81.68} & \textbf{91.38} & \textbf{94.05}\\
			\hline
			\textbf{Slot F1} & en & de & es & fr & hi & ja & pt & tr & zh & \textbf{AVG} \\ 
			\hline
			BiLSTM~\cite{hochreiter1997long} & 75.43 & 15.81 & 34.97 & 33.38 & 5.83 & 4.98 & 43.89 & 9.51 & 27.51 & 27.92\\
			BiLSTM+$\textsc{GL-CLeF}$ & \textbf{87.45} & \textbf{38.40} & \textbf{46.06} & \textbf{46.16} & \textbf{20.28} & \textbf{29.53} & \textbf{59.67} & \textbf{37.25} & \textbf{42.48} & \textbf{45.25}\\
			\hdashline
			XLM-R~\cite{ conneau-etal-2020-unsupervised} & 94.58 & 72.35 & 76.72 & 71.81 & 60.51 & 9.31 & 70.08 & 45.21 & 13.44 & 57.38\\
			XLM-R+$\textsc{GL-CLeF}$ & \textbf{95.88} & \textbf{84.91} & \textbf{82.47} & \textbf{80.99} & \textbf{61.11} & \textbf{55.57} & \textbf{77.27} & \textbf{54.55} & \textbf{80.50} & \textbf{74.81}\\
			\hline
			\textbf{Overall Accuracy} & en & de & es & fr & hi & ja & pt & tr & zh & \textbf{AVG} \\ 
			\hline
			BiLSTM~\cite{hochreiter1997long} & 37.06 & 0.78 & 3.08 & 0.63 & 0.22 & 0.00 & 10.20 & 0.00 & 0.03 & 5.80\\
			BiLSTM+$\textsc{GL-CLeF}$ & \textbf{61.37} & \textbf{4.60} & \textbf{9.10} & \textbf{4.30} & \textbf{0.34} & \textbf{2.03} & \textbf{16.82} & \textbf{2.80} & \textbf{2.46} & \textbf{11.53}\\
			\hdashline
			XLM-R~\cite{ conneau-etal-2020-unsupervised} & 87.45  & 43.05 & 42.93  & 43.74  & 19.42  & 5.76  & 40.80  & 9.65  & 6.60  & 33.31\\
			XLM-R+$\textsc{GL-CLeF}$ & \textbf{88.24}  & \textbf{64.91}  & \textbf{53.51}  & \textbf{58.28}  &  \textbf{19.49} & \textbf{13.77}  & \textbf{52.35}  & \textbf{14.55}  & \textbf{52.07}  & \textbf{46.35}\\
			\hline
		\end{tabular}
\end{adjustbox}
	\caption{Experimental results on BiLSTM and XLM-R. 
	} \label{table:XLMRoberta_Effectiveness}
\end{table*}

\paragraph{Answer 2:  \textit{Semantic-level} \texttt{global} intent-slot successfully establishes a semantic connection across languages.}
We further investigate the effect of the \textit{semantic-level} intent-slot CL mechanism when we remove the \texttt{global} intent-slot CL loss (Figure~\ref{fig:ablation}, ``-- GIS'' (Col 3)).
We find the sentence overall performance drops a lot (from 54.09\% to 46.94\%).
Sentence overall metrics require model to capture the semantic information (intent and slots) for queries.
Therefore, we attribute it to the proposed \textit{semantic-level} \texttt{global} intent-slot CL.  As it successfully establishes semantic connection across languages, it boosts overall accuracy.

\paragraph{Answer 3: Contribution from \texttt{local} intent CL and slot CL module are complementary.}
We explore whether \texttt{local} intent CL and slot CL module are complementary.
By removing all the \texttt{Local} CL modules (including \textit{sentence-level} \texttt{local} intent CL and \textit{token-level} \texttt{local} slot CL), results are shown in  Figure~\ref{fig:ablation}  (--Local Col 4). 
We find that the experiments are lowest compared with only removing any single \texttt{local} CL module, which demonstrates the designed two \texttt{local} CL module works orthogonally.
\pgfplotstableread[row sep=\\,col sep=&]{
    Model & Intent Accuracy. & Slot F1 & Overall Acc. \\
    {-- LI}   & 88.43 & 78.00 & 48.10 \\
    {-- LS}   & 91.80 & 77.56 & 49.56 \\
    {-- Local} & 86.06 & 73.46 & 42.53 \\
    {-- GIS} & 85.33 & 77.32 & 46.94 \\
    {{\scshape GL-CLeF}} & 91.95 & 80.00 & 54.09 \\
    }\mydata

\pgfplotsset{every axis/.append style={
                    label style={font=\small},
                    tick label style={font=\small} 
                    }}
\begin{figure}[t]
    \centering
    \begin{tikzpicture}[scale=0.45]
        \begin{axis}[
                ybar=8pt,
                bar width=0.7cm,
                enlarge x limits=0.2,
                width= 1.1 \textwidth,
                height= 0.8\textwidth,
                legend style={at={(0.5,1)},
                anchor=north,legend columns=-1},
                symbolic x coords=
                 {{-- LI}, {-- LS}, {-- GIS}, {-- Local}, {\scshape GL-CLeF}},
                  xtick=data,
                nodes near coords,
                every node near coord/.append style={font=\normalsize},
                nodes near coords align={vertical},
                ymin=40,ymax=100,
            ]
            \addplot table[x=Model,y=Intent Accuracy.]{\mydata};
            \addplot [draw=black!70!white,fill=black!25!white] table[x=Model,y=Slot F1]{\mydata};
            \addplot table[x=Model,y=Overall Acc.]{\mydata};
            \legend{{Intent Accuracy}, {Slot F1}, {Overall Accuracy}}
        \end{axis}
    \end{tikzpicture}
\caption{Ablation experiments. y-axis denotes the performance score. ``LI'', ``LS'' and ``GIS''  denote \texttt{Local} Intent CL, \texttt{Local} Slot CL, and \texttt{Global} Intent-Slot CL, respectively; ``-Local'' represents removing both ``LI'' and ``LS'' module.}\label{fig:ablation}
\end{figure}
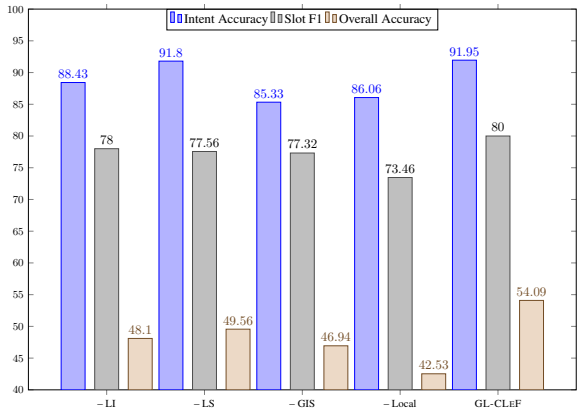

\begin{figure}[t]
	\centering
	\includegraphics[width=0.45\textwidth]{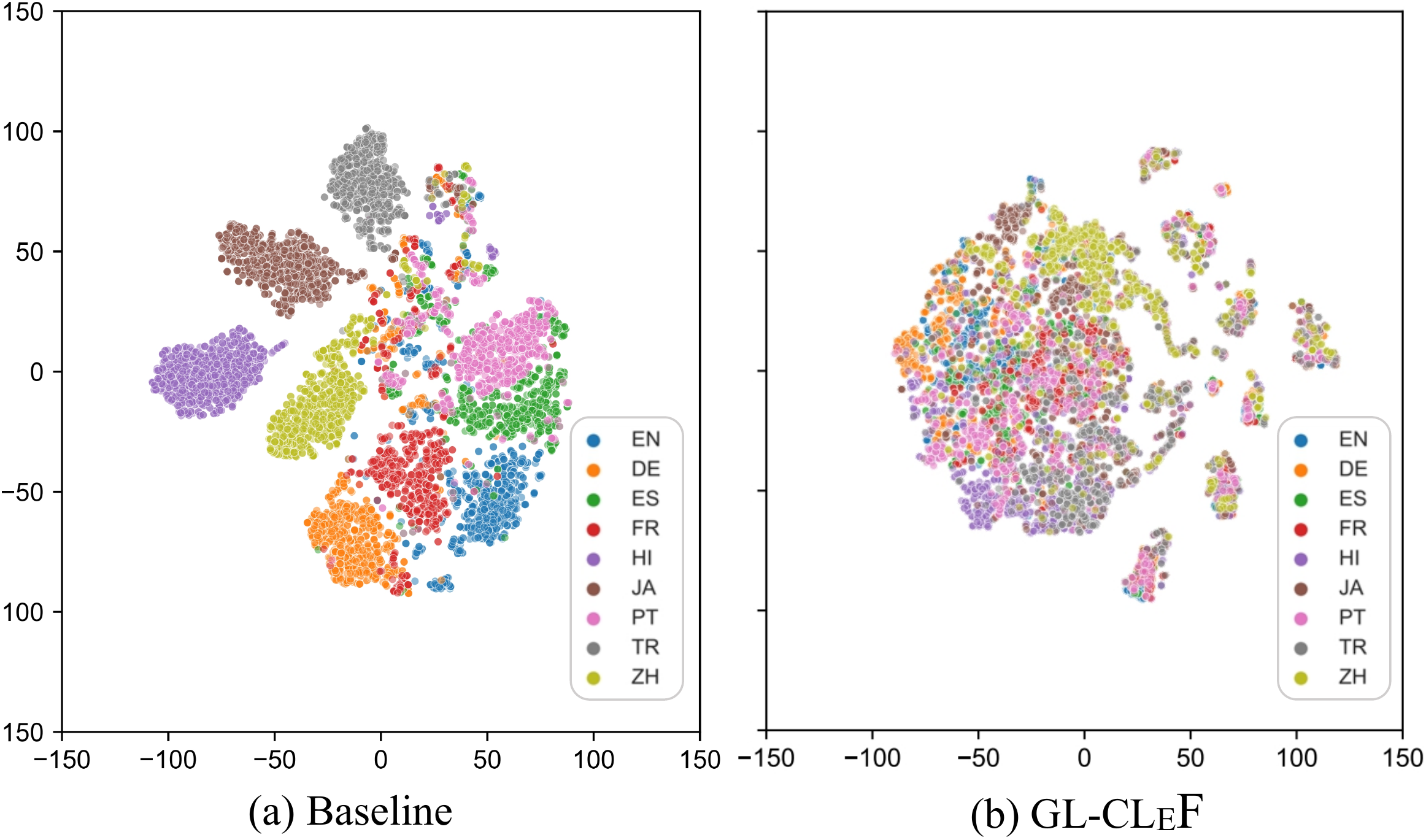}
	\caption{t-SNE visualization of sentence vectors from
(a) mBERT and (b) \textsc{GL-CLeF}. Different colors represents different languages.
}
	\label{fig:visualize}
\end{figure}

\paragraph{Answer 4: \textsc{GL-CLeF} pulls similar representations across languages closer.}
We choose test set and use representations of \texttt{[CLS]} of each sentence for visualization.
Figure~\ref{fig:visualize} (a, LHS) shows the t-SNE visualization of the \texttt{mBERT} output, where we observe that there very little overlap between different languages, which shows that the distance of the representations of different languages are distant. In contrast, the \texttt{\textsc{GL-CLeF}} representations (b, RHS) fine-tuned model in different languages are closer and largely overlap with each other. 
The stark contrast between the figures demonstrates that \texttt{\textsc{GL-CLeF}} successfully aligns representations of different languages.

\paragraph{Answer 5: Contributions from contrastive learning and pre-trained model use are complementary.}
To verify the contribution from \textsc{GL-CLeF} is still effective when used in conjunction with other strong pre-trained models, we perform experiments with \texttt{XLM-R}~\cite{conneau-etal-2020-unsupervised}. \texttt{XLM-R} demonstrates significant gains for a wide range of cross-lingual tasks. 
From the results in Table~\ref{table:XLMRoberta_Effectiveness}, we find \textsc{GL-CLeF} enhances \texttt{XLM-R}'s performance, demonstrating that contributions from the two are complementary.
This also indicates that \textsc{CL-CLeF} is model-agnostic, hinting that \textsc{GL-CLeF} may be applied to other pre-trained models.

\begin{figure}[t]
	\centering
	\includegraphics[width=0.45\textwidth]{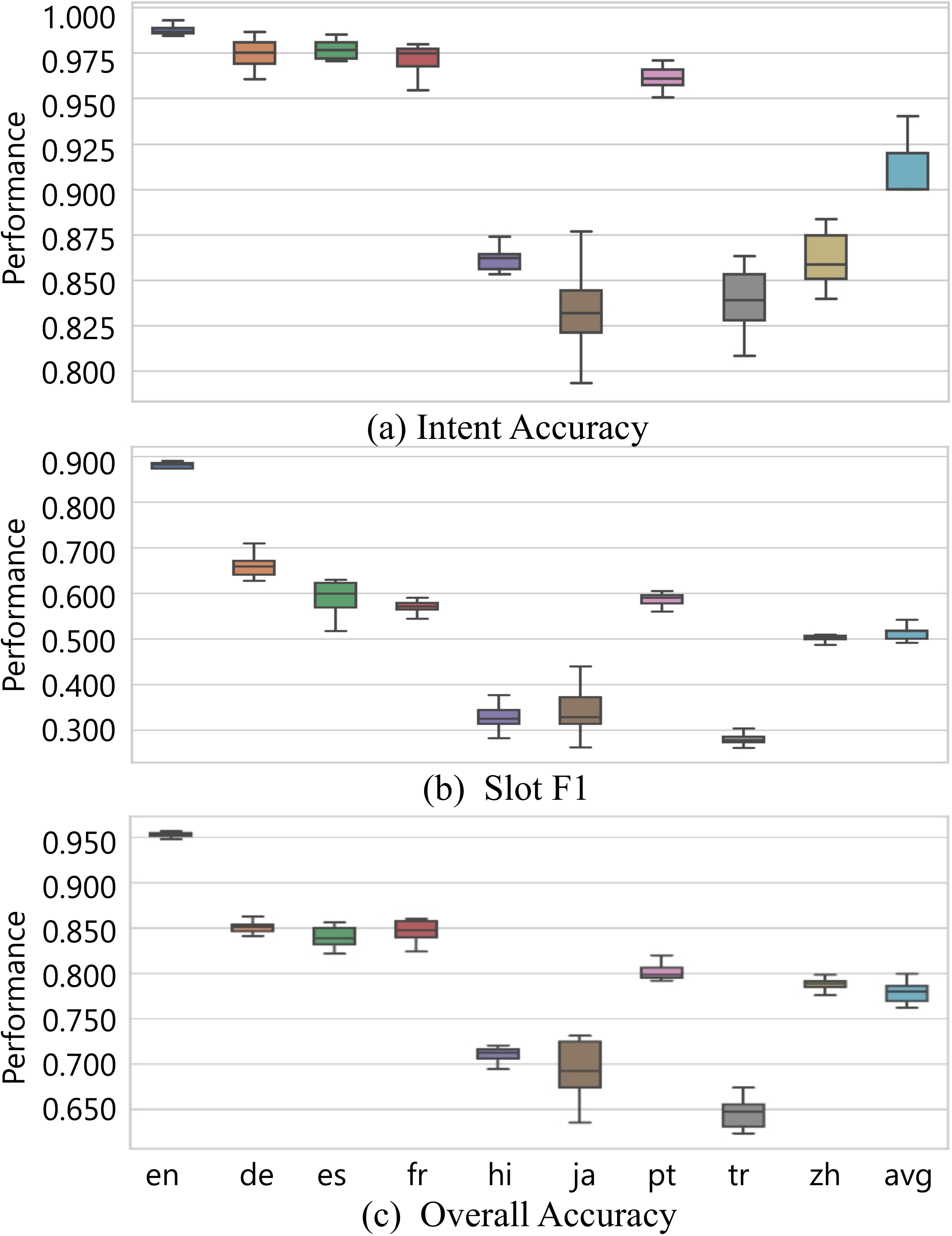}
	\caption{Performance distribution boxplots for each model over 15 random seeds.}
	\label{fig:robust}
\end{figure}
\paragraph{Answer 6: \textsc{GL-CLeF} still obtains gains over BiLSTM.}
A natural question that arises is whether \texttt{\textsc{GL-CLeF}} is effective for non pre-trained models, in addition to transformers. To answer the question, we replace \texttt{mBERT} with \texttt{BiLSTM}, keeping other components unchanged. The results are shown in Table~\ref{table:XLMRoberta_Effectiveness}. We can see that \textsc{GL-CLeF} outperforms \texttt{BiLSTM} in all metrics, further demonstrating that \texttt{\textsc{GL-CLeF}} is not only effective over \texttt{mBERT} but also ports to general encoders for both pre-trained models and non pre-trained models.

\paragraph{Answer 7: \textsc{GL-CLeF} is robust.}
It is worth noting that words in the source language can have multiple translations in the target language.
We follow \citet{ijcai2020-533} to randomly choose any of
the multiple translations as the replacement target language word.  Their work verified that random selection effective method~\cite{ijcai2020-533}. 
A natural question that arises is whether \textsc{GL-CLeF} is robust over different translation selections. 
To answer the question, we choose 15 different seeds to perform experiment and obtain the standard deviation, which we take as an indicator of the stability and robustness of models’ performance.
Results in Figure~\ref{fig:robust} shows
a lower standard deviation on each metric, indicating our model is robust to different translation.  Finding and using the absolutely correct contextual word-to-word translation is an interesting direction to be explored in the future.

%% file: related-work.tex
\paragraph{Traditional Spoken Language Understanding.}
Since slot filling and intent detection are two correlated tasks, traditional SLU approaches mainly explore a joint model for capturing shared knowledge across the two tasks.
Specifically, \citet{zhang2016joint, liu2016attentionbased, liu-lane-2016-joint, hakkani2016multi} consider an implicit joint mechanism using a multi-task framework by sharing an encoder for both tasks.
\citet{goo-etal-2018-slot, li-etal-2018-self, qin-etal-2019-stack}  consider explicitly leveraging intent detection information to guide slot filling.
\citet{wang-etal-2018-bi, haihong2019novel, zhang2020graph, qin2021co} use a bi-directional connection between slot filling and intent detection.

\paragraph{Zero-shot Cross-lingual Spoken Language Understanding.}
Traditional SLU has largely been limited to high-resource languages.  
To solve this problem, zero-shot cross-lingual SLU has gained increasing attention.
Recently, cross-lingual contextualized embeddings have achieved promising results (e.g., mBERT~\cite{devlin-etal-2019-bert}). Many works target improving mBERT at the pre-training stage \cite{conneau2019cross, huang-etal-2019-unicoder,yang2020alternating, DBLP:journals/corr/abs-2007-01852,conneau-etal-2020-unsupervised,xue-etal-2021-mt5,chi-etal-2021-infoxlm,chi2021xlme}.
Compared with their work, our focus is on enhancing mBERT at the fine-tuning stage.

In recent years, related work also considers aligning representations between source and target languages during fine-tuning, eschewing the need for an extra pre-training process.
Specifically, \citet{liu2020attention} propose code-mixing to construct training sentences that consist of both source and target phrases for implicitly ﬁne-tuning mBERT.
\citet{ijcai2020-533} further propose a multi-lingual code-switching data augmentation to better align a source language and all target languages. In contrast to their work, our framework consider aligning similar representation across languages explicitly via a contrastive learning framework.
In addition, in \textsc{GL-CLeF}, we propose a multi-resolution loss to encourage fine-grained knowledge transfer for token-level slot filling.

\paragraph{Contrastive Learning.}
Contrastive learning is now commonplace in NLP tasks.
\citet{DBLP:journals/corr/abs-2012-15466} adopt multiple sentence-level
augmentation strategies to learn a noise-invariant
sentence representation. \citet{DBLP:journals/corr/abs-2005-12766} apply back translation to create augmentations of original sentences for training transformer models.
\citet{wang-etal-2021-cline} propose contrastive learning with semantically negative examples (CLINE) to improve the robustness under semantically adversarial attack.
Inspired by the success of CL, we utilize contrastive learning to explicitly align similar representations across source language and target language.

%% file: Conclusion.tex
We introduced a global--local contrastive learning (CL) framework (\textsc{GL-CLeF}) to explicitly align representations across languages for zero-shot cross-lingual SLU.
Besides, the proposed \texttt{Local} CL module and \texttt{Global} CL module achieves to learn different granularity alignment (i.e., \textit{sentence-level} local intent alignment, \textit{token-level} local slot alignment, \textit{semantic-level} global intent-slot alignment).
Experiments on MultiATIS++ show that \textsc{GL-CLeF} obtains best performance and extensive analysis indicate \textsc{GL-CLeF} successfully pulls closer the representations of similar sentence across languages.

\section{Ethical Considerations}
 Spoken language understanding (SLU) is a core component in task-oriented dialogue system, which becomes sufficiently effective to be deployed in practice.
Recently, SLU has achieved remarkable success, due to the evolution of pre-trained models.
However, most SLU works and applications are English-centric, which makes it hard to generalize to other languages without annotated data.  Our
work focuses on improving zero-shot cross-lingual SLU model that do not need any labeled data for target languages,
which potentially is able to build multilingual SLU models and further promotes the globalization of task-oriented dialog systems.

\section*{Acknowledgements}
We also thank all anonymous reviewers for their constructive comments. 
This work was supported by the National Key R\&D Program of China via grant 2020AAA0106501 and the National Natural Science Foundation of China (NSFC) via grant 61976072 and 62176078.